\documentclass[lnbip]{svmultln}
\usepackage{makeidx}  
\usepackage{fontawesome5}
\usepackage{marvosym}
\newcommand*\myglobe{\includegraphics[height=2ex]{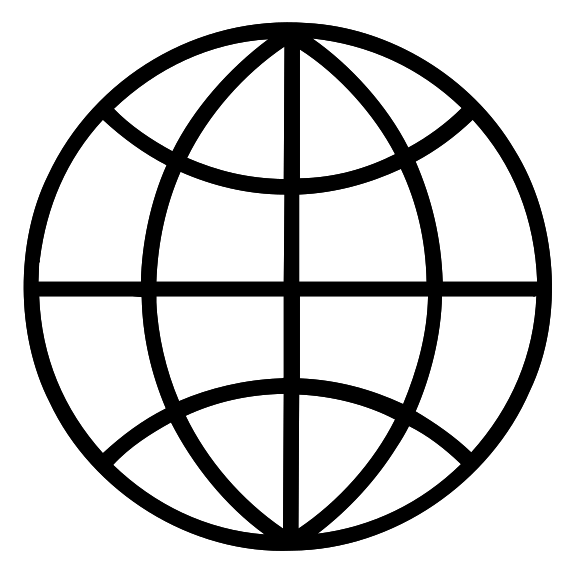}}
\usepackage{geometry}
\usepackage{microtype}
\usepackage[T1]{fontenc}
\usepackage[utf8]{inputenc}
\usepackage{comment}
\usepackage[english]{babel}
\usepackage{algorithmic}
\usepackage[ruled,vlined,linesnumbered]{algorithm2e}
\usepackage{enumitem}
\usepackage{setspace}
\setlist{nosep} 

\usepackage{csquotes}
\usepackage{graphicx}
\usepackage{xcolor}
\usepackage{authblk}
\usepackage[sorting=none,maxbibnames=99,firstinits]{biblatex}
\bibliography{mybib.bib}
\usepackage{float}
\usepackage{url}
\usepackage[normalem]{ulem}
\usepackage{xspace,fancyhdr}
\usepackage{setspace}
\usepackage{amssymb}
\usepackage{amsmath}
\usepackage{amsfonts}
\usepackage[colorinlistoftodos,textwidth=20mm]{todonotes}
\DeclareMathOperator*{\argmax}{arg\,max}

\usepackage{mathtools} 
\usepackage{thmtools}
\usepackage{empheq} 
\usepackage{booktabs}
\usepackage{algorithmic}
\usepackage{multirow}
\usepackage{siunitx}
\usepackage[toc,title,page]{appendix}
\usepackage{hyperref}
\hypersetup{
    colorlinks=true,
    linkcolor=blue,
    filecolor=blue,      
    urlcolor=blue,
    citecolor=red,
    pdftitle={Overleaf Example},
    pdfpagemode=FullScreen,
    }
\usepackage{fullpage}

\usepackage{fancyhdr}

\usepackage{calc}
\usepackage{changepage}
\usepackage{glossaries}
\glsdisablehyper
\usepackage{glossaries}
\makeglossaries
\newacronym{PM}{PM}{Process Mining}
\newacronym{BPM}{BPM}{Business Process Management}
\newacronym{ML}{ML}{Machine Learning}
\newacronym{DM}{DM}{Data Mining}
\newacronym{LHS}{LHS}{Left-Hand-Side}
\newacronym{RHS}{RHS}{Right-Hand-Side}
\newacronym{CNN}{CNN}{Convolutional Neural Network}
\newacronym{DK}{DK}{Deterministically known}
\newacronym{SK}{SK}{Stochastically known}

\newcommand{\avi}[1]{\todo{Avi: #1}}
\usepackage{titling}
\usepackage{blindtext}
\begin{document}
\mainmatter              
\title{Uncertain Process Data with Probabilistic Knowledge:\\ Problem Characterization and Challenges}
\titlerunning{Uncertain process data with probabilistic knowledge}  
%
\author{Izack Cohen\inst{1} \and Avigdor Gal\inst{2}
}
\authorrunning{Cohen and Gal}   
%
%
\institute{
Bar-Ilan University, Ramat-Gan, Israel\\
\Letter~ \texttt{\href{mailto:izack.cohen@biu.ac.il}{izack.cohen@biu.ac.il}} 
~~\myglobe~ \texttt{\url{https://izackcohen.com}}
\and
The Technion - Israel Institute of Technology, Haifa, Israel\\
\Letter~ \texttt{\href{mailto:avigal@technion.ac.il}{avigal@technion.ac.il}}~~ \myglobe ~
\texttt{\url{https://ie.technion.ac.il/~avigal}}
}
\maketitle              
\begin{abstract}        
Motivated by the abundance of uncertain event data from multiple sources including physical devices and sensors, this paper presents the task of relating a stochastic process observation to a process model that can be rendered from a dataset. In contrast to previous research that suggested to transform a stochastically known event log into a less informative uncertain log with upper and lower bounds on activity frequencies, we consider the challenge of accommodating the probabilistic knowledge into conformance checking techniques. Based on a taxonomy that captures the spectrum of conformance checking cases under stochastic process observations, we present three types of challenging cases. The first includes conformance checking of a stochastically known log with respect to a given process model. The second case extends the first to classify a stochastically known log into one of several process models. The third case extends the two previous ones into settings in which process models are only stochastically known. The suggested problem captures the increasingly growing number of applications in which sensors provide probabilistic process information.

\keywords {conformance checking, stochastically known traces, process classification, sensors}
\end{abstract}
\section{Motivation and Problem Description} \label{sec:descr}
Current times are characterized by increasing amounts of event data that are generated from multiple sources including physical devices and sensors. The source of such data may be video clips from social media \cite{sener2018unsupervised}, multiple sources in a smart city (e.g., the \href{https://biu-vf-project.wixsite.com/biuvfproject}{`Green Wall'} project in Tel-Aviv and Nanjing), various medical devices, recording of conversations, and more. The quality of such sources may be low and questionable due to many factors, among them the quality of data capturing devices and quality reduction as part of data processing. The end result of collecting such data into process logs was described in the literature as {\em uncertain sensor data}~\cite{van2016data}. 

In this work, we focus on the problem of managing uncertain process data whenever event data can be characterised in probabilistic terms. As an illustration, consider the use of a machine learning algorithm to detect activities in video clips. Such algorithm typically offers, as a last stage before decision making, a probability distribution over a space of alternatives. The probabilistic information can be utilized to quantify the uncertainty associated with event data, and propagate it to the log to create a stochastic, rather than deterministic, log.

To motivate the problem, consider video cameras as a data source and food preparation as the process domain.
Accordingly, think about a restaurant kitchen that is monitored by video cameras. The cook, who prepares drinks and foods, works according to recipes (i.e., process models). 
We note, in passing, that there 
are multiple supervised food preparation datasets that can be used for process mining research such as, University of Dundee 50 Salads (50Salads) and the Georgia Tech Egocentric Activities (GTEA). 
\begin{figure}[htpb]
	\label{fig:Eggs}  
	\begin{center}	
		\includegraphics[width=0.6\textwidth]{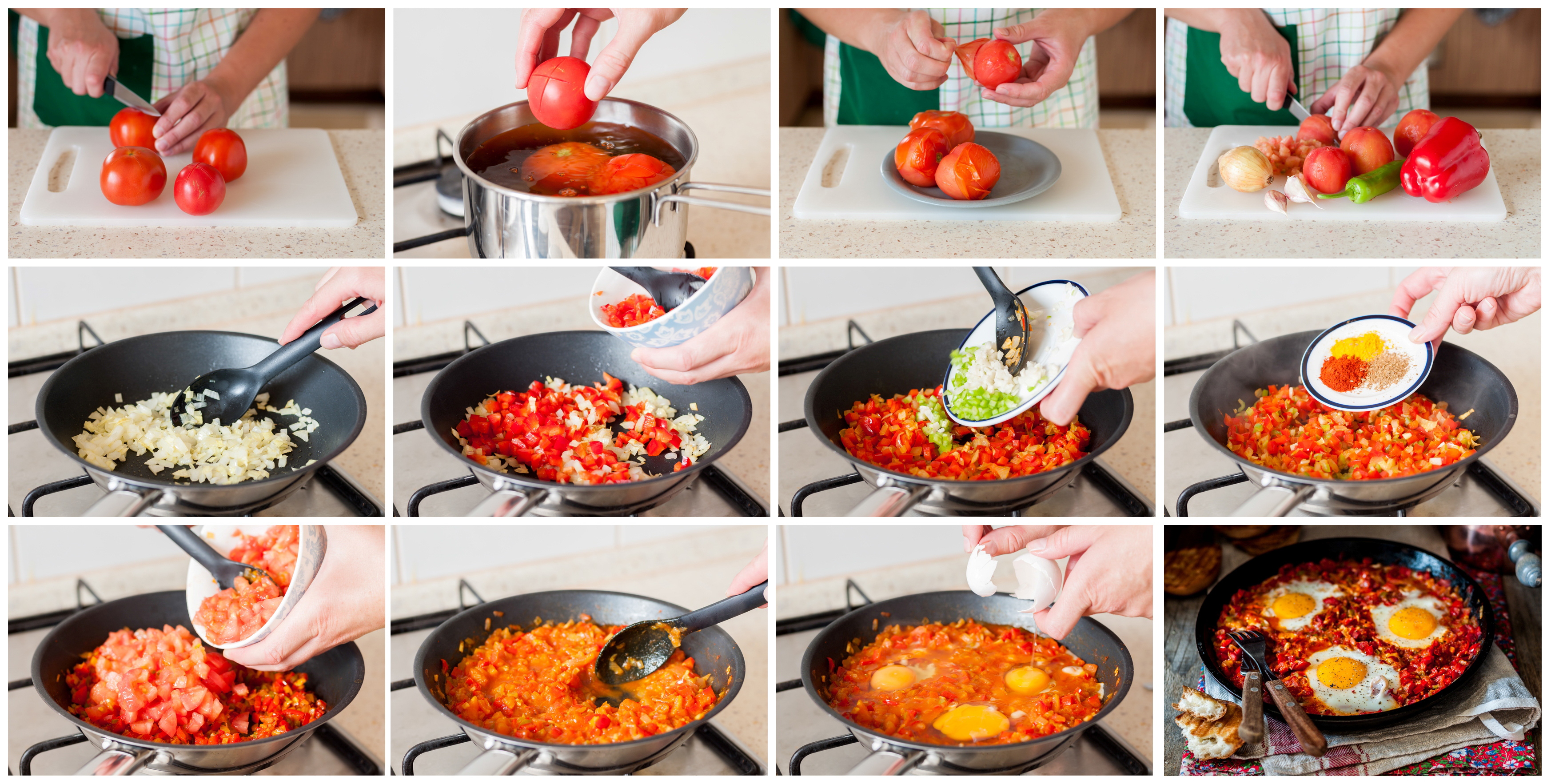}
			\caption{An illustration of a 12-activity food preparation process.}
	\end{center}
	\vskip-.15in
\end{figure}
Given a known (or discovered) set of models (e.g., cookbook recipes or historical supervised datasets), we wish to automatically identify, based on video clips, a prepared dish (e.g., Figure~\ref{fig:Eggs}). Such identification can serve various purposes including conformance of a dish preparation with its recipe, informing diners regarding expected dish arrival time or performance improvement by identification of bottlenecks in the kitchen.

The challenge follows from the fact that the predicted trace, which is the result of data processing and learning techniques, is probabilistic (e.g., a softmax layer of a neural network). The matrix below represents a stochastic trace prediction for 12 events $(e_1,\dots,e_{12})$ and $n$ possible activity classes $(a_1,\dots,a_n)$:
\[
\begin{tabular}{ c } 
 $a_1$ \\ 
 $a_2$ \\
 $\vdots$ \\  
 $a_{n-1}$ \\ 
 $a_n$    
\end{tabular}
  \stackrel{\mbox{$e_1~~~~~e_2~~~\cdots~~~~e_{11}~~~~~~e_{12}$}}{%
    \begin{bmatrix}
p_{1,1} & p_{1,2} & \cdots & p_{1,11} & p_{1,12}\\
p_{2,1} & p_{2,2} & \cdots & p_{2,11} & p_{2,12}\\
\vdots & \vdots & \vdots\vdots\vdots & \vdots & \vdots\\
p_{n-1,1} & p_{n-1,2} & \cdots & p_{n-1,11} & p_{n-1,12}\\
p_{n,1} & p_{n,2} & \cdots & p_{n,11} & p_{n,12}
\end{bmatrix}.%
  }\ 
\]
Assuming a complementary background activity, then for all events $j$, we can generate a probability space such that $ \sum_{i=1}^n p_{i,j} = 1. $
In practice, we expect the problem to battle a large number of events, much larger than the number of events in the toy datasets (e.g., 12 for Figure~\ref{fig:Eggs}). The number of events depends on the length of the overall process and the sampling resolution, which may result in a large number of video frames.
Also, whenever sampling is performed at a predetermined frequency, time points should be grouped into higher level activities. Therefore, the magnitude of the challenge can be understood by the large number of possible traces that  follows from the uncertain trace representation and the fact that to date, no conformance technique was proposed to handle this type of stochastic uncertain traces.   

To jump-start the discussion, we present the related literature in Section~\ref{sec:lit}, followed by a taxonomy to characterise the problem dimensions (Section~\ref{sec:Taxonomy}), where we also present the challenge in more details.

\section{Related Literature} \label{sec:lit}
We first review the scarce research about \gls{PM} with uncertain data. Then, we add context to the use-case on which we focus -- video cameras -- by mentioning related computer vision studies.

Data quality and uncertainty in the context of \gls{PM} have been studied from different perspectives. Several studies focused on data quality and imperfection aspects \cite{suriadi2017event, wang2015cleaning, conforti2016filtering, sani2017improving, van2018filtering, conforti2018timestamp}. These studies have dealt with data quality issues such as wrong event timestamps, a missing linkage between an event and its case-id, and a different description for the same activity. The methodological focus was on preprocessing methods for filtering the affected data or repairing the data values.  

\citeauthor{ceylan2021open} \cite{ceylan2021open} noted that extracting structured data from knowledge (e.g., images, text and speech) by applying statistical techniques such as machine learning models, necessarily creates uncertain data that include probability values for predicted classes. Therefore, data uncertainty has been researched in the context of probabilistic databases and data mining applications, where attributes and/or records are associated with probability distribution functions (e.g., \cite{suciu2011probabilistic}). 

Research about performing \gls{PM} tasks with uncertain data emerged during the last couple of years, by a small group of researchers that included \citeauthor{pegoraro2020conformance} and their associates. \citeauthor{pegoraro2020conformance} \cite{pegoraro2020conformance} and \citeauthor{pegoraro2019mining} \cite{pegoraro2019mining} introduced a taxonomy of uncertain event logs and models. They defined two types of uncertainty: \textit{strong uncertainty} and \textit {weak uncertainty}; strong uncertainty refers to  unknown probability distribution values for attribute values  while weak uncertainty assumes complete probabilistic knowledge (i.e., a probability distribution). The authors suggested a conformance checking technique for a strong uncertainty setting and a way to transform a weakly uncertain log into a strongly uncertain one. Such transformation, however, results in an information loss.  \citeauthor{pegoraro2019discovering} \cite{pegoraro2019discovering}  suggested a discovery technique over strongly uncertain logs. Uncertain activities and arcs in the discovered model can be filtered based on upper and lower bounds on the occurrence frequency of activities and direct relationships between activities. Another stream of research focuses on developing efficient ways to construct behaviour graphs from strongly uncertain longs. These graphs, which consist of a graphical representation of precedence relationships among events \cite{pegoraro2020efficient,pegoraro2020efficientspaceandtime}, form the foundations for model discovery by using methods based on directly-follows relationships such as the Inductive miner \cite{pegoraro2019discovering}. 

Computer vision literature typically refers to process discovery as  `complex activity recognition', which similarly to \gls{PM}, consists of a set of sensor-detected temporally-linked lower-level events. Thus, computer vision based process discovery is dependent upon automatically recognizing \textit{simple} activities from which the process is composed such as `walking', `jumping', `meeting' and the temporal links between them; and this task poses a challenge for current machine learning techniques  \cite{zhang2019comprehensive, ma2020sf}.

In this paper, we focus on the challenge of weakly uncertain logs that were only mentioned casually in past research \cite{pegoraro2020conformance}. We believe that weakly uncertain settings, which are increasingly common in many applications, need (and can) be explicitly dealt with. While data uncertainty may extend across several attributes we focus on the control-flow aspect which implicitly accommodates the aspect of time.

\section{Taxonomy, Challenges and Initial Solution Ideas} \label{sec:Taxonomy} 
 
 To characterize environments of interest, we define two terms, namely \gls{DK} and \gls{SK}. The former refers to a process model or an event log that are given and deterministic (e.g., a supervised dataset of video movies). The latter refers to a known probability distribution of event attribute values in an observed event log (e.g., to a testing dataset of video movies). Accordingly, for a \gls{SK} trace within a dataset, the probability distribution of each event to be classified as one of the possible activities is known. 

 \begin{table}[ht]
	\centering
	\bgroup
\def\arraystretch{1.3}
\begin{tabular}{l p{2cm}  p{2cm} p{2cm} p{2cm}}
				\midrule
				\hskip.6in \textbf{Model (Dataset)} $\rightarrow$ & \multicolumn{2}{l}{Single process}  & \multicolumn{2}{l}{Multiple processes}\\
				$\downarrow$ \textbf{Observation (Log)} & \gls{DK} & \gls{SK} & \gls{DK}& \gls{SK}\\ \midrule
				Deterministically Known (\gls{DK})    & 1 & 2 & 3 & 4\\
				   \midrule
				Stochastically Known (\gls{SK}) &  5 & 6   &7 & 8\\
				       \midrule
		\end{tabular}
		\egroup
		\caption{Eight cases according to the characteristics of the process and observed log}
		\label{tab:ProcLog}
\end{table}

Table~\ref{tab:ProcLog} accommodates the spectrum of conformance checking using the \gls{SK} term. 
Case 1 is the standard conformance checking where process realizations are compared to a process model. Case 3 uses conformance for classification where several processes are given and the observation is classified to the process model with which it conforms the most. Thus, conformance checking is performed with respect to each of the known processes. Cases 5 and 7 relate to weakly uncertain observed logs. Case 5 may represent a setting in which one wants to check, for example, the conformance of a surgical procedure with its model (e.g., for educating surgeons or debriefing purposes). Such a case poses the challenge of developing a conformance technique that explicitly accommodates the probabilistic information. In such a case, an example observation may be modeled by the following probability matrix:
\[
\begin{tabular}{ c }
 $a$ \\ 
 $b$ \\
 $c$ \\  
 $d$    
\end{tabular}
  \stackrel{\mbox{$e_1~~~e_2~~~e_3~~~e_4$}}{%
    \begin{bmatrix}
0.50 & 0.30 & 0.10 & 0.20\\
0.30 & 0.60 & 0.10 & 0.20\\
0.20 & 0.05 & 0.20 & 0.31\\
0.00 & 0.05 & 0.60 & 0.29
\end{bmatrix},%
  }\ 
\]
where rows correspond to activities (e.g., $a$-$d$), columns to timestamps (e.g., $e_1$-$e_4$), and entries represent the probability of an activity to occur in a  time point.
The matrix can be the outcome of a softmax layer of a neural network; the probabilities associated with the first event $e_1$, for example, are $p(a)=0.50,\,p(b)=0.30,\,p(c)=0.20$, and $p(d)=0.00$. We note that the presentation implicitly captures time uncertainty; for example, consider events that represent the sensor sampling time -- that is, $e_1,e_2,\dots$ represent time moments in which probabilistic information about activities was gathered. Thus, an activity duration may be represented by a time interval between events, e.g.  $t(e_j)-t(e_i),\ e_j \succ e_i,$ with some probability. 

In Case 7, an observed process needs to be classified into one of the process models using a conformance measure. A representative use-case may include a dataset of food preparation \gls{DK} models (e.g., latte, tea, scrambled eggs, and cheese sandwich) and a \gls{SK} log based on a video recorded dish preparation that needs to be automatically classified as one of the models. In such a case, we suggest conformance checking of the observation with respect to each of the models---the best conforming model is selected as the prepared dish. The challenge is to develop the conformance checking procedures for the probabilistic setting.

In Cases 2,4,6 and 8, the models are \gls{SK}. Such settings may arise when creating a fully supervised dataset is too costly. A natural way to discover the models is to apply neural network techniques on videos of known dishes, which would result in a \gls{SK} trace for each historical video with a deterministically known label (i.e., the dish name is known). Cases 6 and 8 in which both models and the log are \gls{SK}, are the most challenging. We expect that it would be extremely hard to distinguish between two types of stochasticity. The first reflects variations across process realizations (e.g., in $60\%$ of the realizations $a\rightarrow b$ and in the rest $a\rightarrow c$) and the second type reflects quality discrepancies induced by sensors and statistical data processing techniques (e.g., the second event is $b$ with probability of $0.6$ or $c$ with probability of $0.4$).

To recapitulate, we introduced a set of challenging conformance and classification problems one needs to address when logs use uncertain data that were generated by devices, sensors and data processing algorithms. The difference with respect to related work is both in the taxonomy and the explicit way in which we model and deal with uncertainty. Modeling and solution methods will require extending conformance methods (e.g., alignments) or developing new ones based on probabilistic measures (e.g., Frobenius norm, Cross-entropy) and new cost structures.

\vspace{-10pt}
\setstretch{1}
\printbibliography

@article{ceylan2021open,
  title={Open-world probabilistic databases: Semantics, algorithms, complexity},
  author={Ceylan, {\.I}smail {\.I}lkan and Darwiche, Adnan and van den Broeck, Guy},
  journal={Artificial Intelligence},
  volume={295},
  pages={103474},
  year={2021},
  publisher={Elsevier}
}

@article{conforti2016filtering,
  title={Filtering out infrequent behavior from business process event logs},
  author={Conforti, Raffaele and La Rosa, Marcello and ter Hofstede, Arthur HM},
  journal={IEEE Transactions on Knowledge and Data Engineering},
  volume={29},
  number={2},
  pages={300--314},
  year={2016},
  publisher={IEEE}
}

@article{conforti2018timestamp,
  title={Timestamp repair for business process event logs},
  author={Conforti, Raffaele and La Rosa, Marcello and ter Hofstede, Arthur HM},
  journal={Preprint available at https://minerva-access. unimelb. edu. au/handle/11343/209011},
  year={2018}
}

@inproceedings{ma2020sf,
  title={SF-Net: Single-frame supervision for temporal action localization},
  author={Ma, Fan and Zhu, Linchao and Yang, Yi and Zha, Shengxin and Kundu, Gourab and Feiszli, Matt and Shou, Zheng},
  booktitle={European Conference on Computer Vision},
  pages={420--437},
  year={2020},
  organization={Springer}
}

@article{pegoraro2020conformance,
  title={Conformance Checking over Uncertain Event Data},
  author={Pegoraro, Marco and Uysal, Merih Seran and van der Aalst, Wil},
  journal={ArXiv Preprint ArXiv:2009.14452},
  year={2020}
}

@inproceedings{pegoraro2019discovering,
  title={Discovering process models from uncertain event data},
  author={Pegoraro, Marco and Uysal, Merih Seran and van der Aalst, Wil},
  booktitle={International Conference on Business Process Management},
  pages={238--249},
  year={2019},
  organization={Springer}
}

@inproceedings{pegoraro2019mining,
  title={Mining uncertain event data in process mining},
  author={Pegoraro, Marco and van der Aalst, Wil},
  booktitle={2019 International Conference on Process Mining (ICPM)},
  pages={89--96},
  year={2019},
  organization={IEEE}
}

@inproceedings{pegoraro2020efficient,
  title={Efficient construction of behavior graphs for uncertain event data},
  author={Pegoraro, Marco and Uysal, Merih Seran and van der Aalst, Wil},
  booktitle={International Conference on Business Information Systems},
  pages={76--88},
  year={2020},
  organization={Springer}
}

@article{pegoraro2020efficientspaceandtime,
  title={Efficient Time and Space Representation of Uncertain Event Data},
  author={Pegoraro, Marco and Uysal, Merih Seran and van der Aalst, Wil},
  journal={Algorithms},
  volume={13},
  number={11},
  pages={285},
  year={2020},
  publisher={Multidisciplinary Digital Publishing Institute}
}

@inproceedings{sani2017improving,
  title={Improving process discovery results by filtering outliers using conditional behavioural probabilities},
  author={Sani, Mohammadreza Fani and van Zelst, Sebastiaan J and van der Aalst, Wil},
  booktitle={International Conference on Business Process Management},
  pages={216--229},
  year={2017},
  organization={Springer}
}

@inproceedings{sener2018unsupervised,
  title={Unsupervised learning and segmentation of complex activities from video},
  author={Sener, Fadime and Yao, Angela},
  booktitle={Proceedings of the IEEE Conference on Computer Vision and Pattern Recognition},
  pages={8368--8376},
  year={2018}
}

@article{suriadi2017event,
  title={Event log imperfection patterns for process mining: Towards a systematic approach to cleaning event logs},
  author={Suriadi and Andrews, Robert and Ter Hofstede, Arthur HM and Wynn, Moe Thandar},
  journal={Information Systems},
  volume={64},
  pages={132--150},
  year={2017},
  publisher={Elsevier}
}

@article{suciu2011probabilistic,
  title={Probabilistic databases, synthesis lectures on data management},
  author={Suciu, Dan and Olteanu, Dan and R{\'e}, Christopher and Koch, Christoph},
  journal={Morgan \& Claypool},
  year={2011}
}

@incollection{van2016data,
  title={Data science in action},
  author={van der Aalst, Wil},
  booktitle={Process Mining},
  pages={3--23},
  year={2016},
  publisher={Springer}
}

@inproceedings{van2018filtering,
  title={Filtering spurious events from event streams of business processes},
  author={van Zelst, Sebastiaan J and Sani, Mohammadreza Fani and Ostovar, Alireza and Conforti, Raffaele and La Rosa, Marcello},
  booktitle={International Conference on Advanced Information Systems Engineering},
  pages={35--52},
  year={2018},
  organization={Springer}
}

@inproceedings{wang2015cleaning,
  title={Cleaning structured event logs: A graph repair approach},
  author={Wang, Jianmin and Song, Shaoxu and Lin, Xuemin and Zhu, Xiaochen and Pei, Jian},
  booktitle={2015 IEEE 31st International Conference on Data Engineering},
  pages={30--41},
  year={2015},
  organization={IEEE}
}

@article{zhang2019comprehensive,
  title={A comprehensive survey of vision-based human action recognition methods},
  author={Zhang, Hong-Bo and Zhang, Yi-Xiang and Zhong, Bineng and Lei, Qing and Yang, Lijie and Du, Ji-Xiang and Chen, Duan-Sheng},
  journal={Sensors},
  volume={19},
  number={5},
  pages={1005},
  year={2019},
  publisher={Multidisciplinary Digital Publishing Institute}
}
\end{document}